\documentclass{article}

\usepackage[english]{babel}
\usepackage{graphicx} % Required for inserting images
\usepackage{authblk}
\usepackage{orcidlink} % package for ORCID icon
\usepackage{caption}
\usepackage{float}
\usepackage{doi}
\usepackage{svg}
\usepackage{amsmath}
\usepackage{amsfonts}
\usepackage{subcaption}

\providecommand{\keywords}[1]{\textbf{\textit{Keywords}} #1}

\title{Weakly Supervised Pneumonia Localization from Chest X-Rays Using Deep Neural Network and Grad-CAM Explanations}

\author[1,*]{Kiran Shahi\,\orcidlink{0000-0003-3739-5985}}

\author[2]{Anup Bagale\,\orcidlink{0009-0001-8261-7113}}

\affil[1]{MBS Survey Software LTD., Steyning, United Kingdom}

\affil[2]{Frontline Hospital, Kathmandu, Nepal}

\affil[*]{Corresponding author(s). E-mail(s): ks@mbs-software.co.uk; Contributing authors: bagaleanup1@gmail.com;}

\date{August 2025}

\begin{document}

\maketitle

\begin{abstract}
Chest X-ray imaging is commonly used to diagnose pneumonia, but accurately localizing the pneumonia affected regions typically requires detailed pixel-level annotations, which are costly and time consuming to obtain. To address this limitation, this study proposes a weakly supervised deep learning framework for pneumonia classification and localization using Gradient-weighted Class Activation Mapping (Grad-CAM). Instead of relying on costly pixel-level annotations,  the proposed method utilizes image-level labels to generate clinically meaningful heatmaps that highlight pneumonia affected regions. Furthermore, we evaluate seven pre-trained deep learning models including a Vision Transformer under identical training conditions, using focal loss and patient-wise splits to prevent data leakage. Experimental results suggest that all models achieved high classification accuracy (96–98\%), with ResNet-18 and EfficientNet-B0 showing the best overall performance and MobileNet-V3 providing an efficient lightweight alternative. Grad-CAM heatmap visualizations in this study confirm that the proposed methods focus on clinically relevant lung regions, supporting the use of explainable AI for radiological diagnostics. Overall, this work highlights the potential of weakly supervised, explainable models that enhance the transparency and clinical trust in AI-assisted pneumonia screening.
\end{abstract}


\section{Introduction}
Pneumonia is still a leading cause of morbidity and mortality worldwide, especially among children and elderly individuals. Although chest X-ray imaging is the most common diagnostic tool \cite{BARTOLF2016373}, interpreting chest X-rays can be a challenging task due to easily missed, subtle, and ambiguous lesions. Variability in radiologists’ interpretations and the frequent oversight of small abnormalities can make consistent diagnosis difficult \cite{ROPP2015277}. These limitations highlight the need for reliable solutions. As artificial intelligence systems continue to mature, deep learning-based methods offer strong potential for the accurate and efficient detection of pneumonia. 

Prior research has demonstrated the state-of-the-art capabilities of Convolutional Neural Networks (CNNs) and Vision Transformers (ViT) in various medical image analysis tasks, including pneumonia detection from chest X-rays\cite{KERMANY2018}. However, most solutions operate in a "black-box" approach, offering limited insight into influential regions in the X-ray image that drive their decisions. Since radiologists require transparent, localized explanations to verify model outputs, this lack of interpretability restricts clinical adoption. Moreover, pixel-level annotations, such as segmentation masks or bounding boxes, are necessary for fully supervised localization techniques; however, they are costly and challenging to acquire at scale \cite{LITJENS201760}.

Weakly supervised learning (WSL) techniques offer a practical approach for spatial localization using only image level labels, thereby avoiding the burdens of manual pixel wise annotations. Among various WSL approaches, Gradient-weighted Class Activation Mapping (Grad-CAM) has been widely adopted for visual explanations in radiology, highlighting the most influential regions that contribute to model predictions. Therefore, interpretable heatmaps generated using Grad-CAM enhance interpretability and provide clinicians with intuitive visual cues linking predictions to underlying radiographic features.

This study presents a unified framework for pneumonia classification and Grad-CAM based weakly supervised localization of pneumonia using chest X-rays. We benchmark seven different pre-trained model, such as ResNet-18\cite{He_2016_CVPR}, ResNet50\cite{He_2016_CVPR}, DenseNet121\cite{Huang_2017_CVPR}, EfficientNet-B0\cite{pmlr-v97-tan19a}, MobileNetV2\cite{Sandler_2018_CVPR}, MobileNetV3\cite{Howard_2019_ICCV}, and the transformer-based ViT-B16\cite{dosovitskiy2020vit}, under identical training conditions. This article highlights the potential of explainable and weakly supervised AI methods to narrow the gap between automated image interpretation and practical clinical decision-making.

The main contributions of this paper are as follows:
\begin{itemize}
    \item We evaluate a Chest X-Rays dataset \cite{KERMANY2018} with strict patient level split to prevent the data leakage.
    \item We benchmark six pretrained CNN architectures and a Vision Transformer backbones under identical training and evaluation settings.
    \item We integrate Grad-CAM and token activation visualization to produce radiologically meaningful heatmaps aligned with lung regions, offering interpretable AI insights for clinicians.
    \item We identify MobileNet-V3 as an optimal trade off between accuracy and computational cost, supporting real-time, edge and mobile health application.
\end{itemize}

The remainder of the paper is organised as follows. Section 2 reviews previous studies on pneumonia detection, weakly supervised learning and the use of explainability in pneumonia localisation. Section 3 describes the methods and neural architectures employed in the experiments. Section 4 presents the experimental setup, datasets, and analysis of the results. Finally, Section 5 provides the conclusion and future work.

\section{Related Work}
In 2018, Kermany et al. \cite{KERMANY2018} introduced a large chest X-ray dataset dedicated to pneumonia detection, which opened new opportunities for researchers in medical image analysis. Early research on pneumonia detection primarily relied on supervised learning methods and focused mainly on pneumonia classification. For instance, Tilve et al. \cite{Tilve_2020} benchmarked pneumonia detection using both traditional machine learning techniques, such as k-nearest neighbors (KNN), and modern convolutional neural network (CNN) approaches, demonstrating the superior performance of CNN-based supervised methods. Similarly, Erdem and Aydın \cite{erdem2021detection} further proposed a novel CNN framework with separable blocks and transfer learning for efficient pneumonia detection. Similarly, Zavaleta et al. \cite{Zavaleta_2025} demonstrated that lightweight architectures such as MobileNetV2 achieve a favorable balance between predictive accuracy and computational efficiency. Although these supervised models achieved strong classification performance, they relied heavily on large, manually labeled datasets, making them costly to train and prone to overfitting and poor generalization.

Weakly supervised learning (WSL) has emerged as a promising solution to reduce dependence on expensive, pixel-level annotations required for fully supervised models \cite{martel2020medical}. WSL utilizes incomplete or inexact supervision, such as image-level labels or free-text radiology reports, to enable large-scale model training \cite{tam2020weakly,al2022automatic}. Tam et al. \cite{tam2020weakly} introduced a multimodal framework combining object detection with natural language processing (NLP) for semantically grounded localization. Subsequent work extended these ideas using transformer and generative architectures. Saber et al. \cite{saber2024efficient} proposed a multi-scale transformer with lung segmentation and attention mechanisms, while Keshavamurthy et al. \cite{keshavamurthy2021weakly} developed a GAN-based WSL model for fine-grained pneumonia localization without bounding-box labels. Other CNN-based WSL methods \cite{guo2023diagnosis,odaibo2019detection,muller2024weakly} demonstrated accurate localization using only image-level supervision.

While weakly supervised methods are continuously advancing in pneumonia detection, most studies still rely on complex architectures or domain-specific annotations, which limit reproducibility and clinical deployment. Moreover, few works systematically compare CNN and transformer backbones under identical training and evaluation settings. In addition, due to the black-box nature of AI models, many prior studies remain limited to non-interpretable approaches, creating hesitation toward clinical adoption. This study addresses these gaps by introducing a unified benchmarking framework for weakly supervised pneumonia localization using Grad-CAM across seven pretrained models, emphasizing interpretability, computational efficiency, and clinical relevance.

\section{Methods}
\begin{figure}[t]
  \centerline{\includegraphics[width=\linewidth]{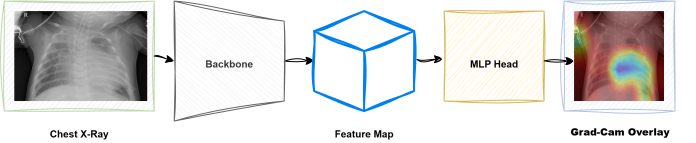}}
  \caption{\label{fig:arch}Model Architecture}
\end{figure}

As illustrated in FIGURE~\ref{fig:arch}, the proposed framework follows a standard deep learning pipeline consisting of dataset preprocessing, feature extraction using pretrained model, training and evaluation of model performance. Furthermore, we compute the class activation maps to create heatmaps that localize the pneumonia affected regions.

\subsection{Dataset}
We used the publicly available Chest X-ray dataset \cite{KERMANY2018}. In this dataset, 1583 X-ray images are in normal class and 4273 are in pneumonia class including both train and test set. In train set 1349 images are in normal class and 3884 images are in pneumonia class. Similarly, in test set 234 images are in normal class and 390 images are in pneumonia class. However, during dataset inspection, we observed that some patients ids were on both training and test sets, which could cause data leakage. To address this issue, we merged the original splits and re-partitioned the dataset at the patient level into training (70\%), validation (15\%), and test (15\%) sets. Each image was resized to 224 x 224 pixels to match the input requirements of ImageNet-pretrained backbones. Since the original images were grayscale, we duplicated the channel three times to create a pseudo-RGB input to match the input shape for pretrained backbones. Further, to enhance generalization, we applied data augmentation including random rotation, horizontal flipping, brightness/contrast adjustment, and Gaussian noise. Dataset splitting was performed at the patient level to prevent data leakage, with 70\% of patients for training, 15\% for validation, and 15\% for testing.

\begin{table}[htbp]
\caption{Dataset distribution after splitting.}
\begin{center}
\begin{tabular}{|c|c|c|}
\hline
\textbf{Subset} & \textbf{NORMAL} & \textbf{PNEUMONIA} \\ \hline
Train (70\%) & 1{,}114 & 2{,}951 \\ \hline
Validation (15\%) & 232 & 653 \\ \hline
Test (15\%) & 237 & 669 \\ \hline
\end{tabular}
\label{dataset_stats}
\end{center}
\end{table}

\begin{figure}[htbp]
\centerline{\includegraphics[width=\linewidth]{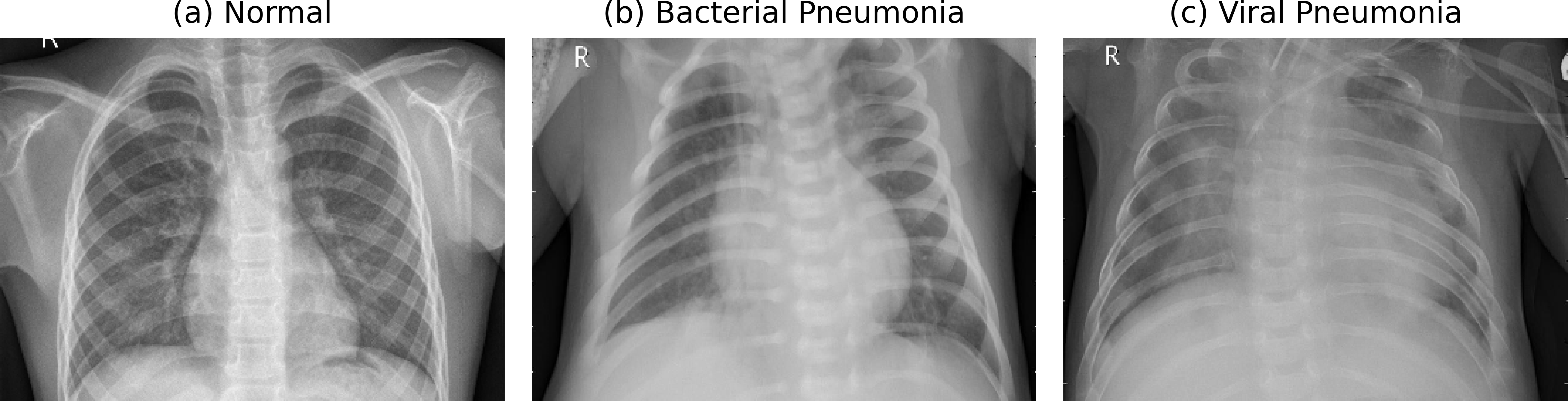}}
\caption{Sample chest X-ray images from the dataset: (a) Normal, (b) Pneumonia (Bacterial), and (c) Pneumonia (Viral).}
\label{fig:dataset}
\end{figure}

\subsection{Model Architectures}

In this study, we evaluated seven different widely used ImageNet pretrained models to explore  different trade-offs between accuracy, efficiency and representational power. These include residual networks, densely connected networks, parameter-efficient scaling methods, mobile optimized networks, and transformer based models.

\begin{itemize}
   \item\textbf{ResNet18 and ResNet50 \cite{He_2016_CVPR}:} Residual networks (ResNet) were introduced by Kaiming He et al. in 2015 to address the vanishing gradient problem by introducing skip connections that enable more stable gradients to flow across layers. ResNet-18, with its 18 layers serves as a lightweight baseline, whereas ResNet-50 with its deeper 50 layers architecture, captures more complex hierarchical features.
   \item\textbf{DenseNet121 \cite{Huang_2017_CVPR}:} In 2016, Gao Huang et al. introduced DenseNet, which improves feature reuse and gradient propagation by connecting each layer to all subsequent layers. This design leads to compact models with fewer parameters while retaining strong representational capacity.
   \item\textbf{EfficientNet-B0 \cite{pmlr-v97-tan19a}:} EfficientNet introduces a compound scaling method that uniformly scales depth, width and resolution using a fixed coefficient. This results in highly parameter-efficient models that achieve strong accuracy with fewer computational resources.
   \item\textbf{MobileNetV2 \cite{Sandler_2018_CVPR} and MobileNetV3 \cite{Howard_2019_ICCV}:} MobileNetV2 and MobileNetV3 are lightweight architectures designed for efficient deployment on mobile and edge devices. MobileNetV2 employs inverted residual blocks with linear bottlenecks to reduce computational cost, while MobileNetV3 further integrates squeeze-and-excitation modules and neural architecture search to improve the latency–accuracy trade-off.
   \item\textbf{ViT-B16 \cite{dosovitskiy2020vit}:} Transformer architectures have become the de facto standard for natural language processing tasks. Building on this success, Alexey Dosovitskiy et al. extended the transformer framework to vision by proposing the Vision Transformer (ViT), which replaces convolutional operations with self-attention and processes images as sequences of non-overlapping patches. In this study, we include the ViT-B/16 model to compare transformer-based architectures with traditional CNNs. ViT-B/16 splits each image into 16×16 pixel patches and processes the resulting sequence using a transformer encoder.
\end{itemize}

All of the above mentioned pretrained models were integrated with a custom classification head consisting of fully connected layers, batch normalization, ReLU activation and dropout layers ensuring fair comparison.

\subsection{Training Procedure}
All models were initialized with ImageNet-pretrained weights to leverage transfer learning. Training was conducted under identical protocols to ensure a fair comparison between backbones.
\begin{itemize}
    \item \textbf{Input preprocessing:} Each image was resized to 224×224 pixels and normalized with ImageNet mean and standard deviation. Since the dataset is grayscale, the channel was duplicated three times to create a pseudo-RGB input to match the input shape of pretrained models.
    \item  \textbf{Data augmentation:} To improve generalization and mitigate overfitting, $\pm 15^{\circ}$ rotations, $\pm5 \%$ affine transformation, 5\% brightness contrast adjustment, CLAHE, gamma correction, Gaussian noise, motion blur, median blur and coarse dropout were applied.
    \item \textbf{Loss functions:} We evaluated three options Cross-Entropy Loss, Weighted Cross-Entropy Loss, and Focal Loss \cite{Lin_2020}. Focal Loss was ultimately chosen as it provided improved handling of the severe class imbalance (404 normal vs. 3692 pneumonia images).

    The Focal Loss is an extension of the standard binary cross-entropy to better handle class imbalance by reducing the relative loss for well-classified class. It introduces a focusing parameter $\gamma$ that down-weights easy samples, allowing the model to concentrate more on hard or misclassified cases.
    
    Mathematically the binary focal loss function is defined as:
    \begin{equation*}
    L(y, \hat{p}) =  - \alpha \, y (1 - \hat{p})^{\gamma} \log(\hat{p}) - (1 - \alpha) (1 - y) \, \hat{p}^{\gamma} \log(1 -\hat{p}) 
    \label{eq:focal_loss}
    \end{equation*}
where $y \in \{0,1\}$ is the ground-truth label and $\hat{p} \in [0,1]$ is the predicted probability for the positive class. The parameter $\gamma$ controls how strongly easy examples are down weighted higher values increase the focus on hard samples while  $\alpha$ balances the importance between positive and negative classes. When $\gamma = 0$, the Focal Loss simplifies to the standard weighted binary cross-entropy loss.
    \item \textbf{Class imbalance strategies:} In addition to Focal Loss, we applied random over-sampling of minority class during training. This ensured that each mini-batch was more balanced and prevented the model from being biased toward the pneumonia class.
    \item \textbf{Optimizer and hyperparameters:} All models were trained using the Adam optimizer with a learning rate of $1\times10^{-4}$ and a weight decay of $1\times10^{-4}$.Training was performed with a batch size of 32 for up to 10 epochs, with early stopping applied to prevent overfitting.
    \item \textbf{Model checkpoint and early stopping:} For each training loop, the best model checkpoint was selected according to validation accuracy and ROC-AUC score. Early stopping was employed to mitigate overfitting when no improvement in validation performance was observed for three consecutive epochs.
    \item \textbf{Evaluation:} After each training loop, each model was evaluated on an independent test set of chest X-ray images. To ensure fairness and reproducibility, we assessed our methods using standard evaluation metrics, including accuracy, ROC-AUC, PR-AUC, and the best F1-score. Each evaluation metric is explained in the following section with its mathematical formulation.
\end{itemize}

\subsection{Performance Evaluation Metrics}
To systematically assess model performance, we employed a set of evaluation metrics designed to measure both classification accuracy and clinical relevance in class-imbalanced conditions. To formulate evaluation metrics mathematically, let us assume TP, TN, FP, and FN represents true positives, true negatives, false positives, and false negatives, respectively.

\textbf{Accuracy}:  
\[
\text{Accuracy} = \frac{\text{TP} + \text{TN}}{\text{TP} + \text{TN} + \text{FP} + \text{FN}}
\]

Accuracy measures the proportion of correct predictions both true pneumonia cases (TP) and true normal cases (TN) out of all predictions. However, accuracy can be misleading in imbalanced datasets because it may overestimate performance by favoring the majority class. Therefore, we evaluate our models using additional class-imbalance–aware metrics.

\textbf{Precision (Positive Predictive Value)}:  
\[
\text{Precision} = \frac{\text{TP}}{\text{TP} + \text{FP}}
\]

Precision quantifies how many of model's positive predictions were actually true positive.

\textbf{Recall (Sensitivity / True Positive Rate)}:  
\[
\text{Recall} = \frac{\text{TP}}{\text{TP} + \text{FN}}
\]

Similarly, recall measures how many of the actual positive cases (pneumonia) the model correctly identifies. A high recall indicates that the model misses very few pneumonia cases. This is clinically important because false negatives failing to detect pneumonia can lead to potentially serious consequences.

\textbf{Specificity (True Negative Rate)}:  
\[
\text{Specificity} = \frac{\text{TN}}{\text{TN} + \text{FP}}
\]
Specificity measures how well the model identifies normal cases. A high specificity indicates that few normal X-rays are incorrectly predicted as pneumonia.

\textbf{F1-score}:  
\[
F1 = 2 \cdot \frac{\text{Precision} \cdot \text{Recall}}{\text{Precision} + \text{Recall}}
\]  
The F1-score is the harmonic mean of precision and recall. In this study, we report the best F1-score obtained across all classification thresholds.

\textbf{ROC-AUC (Receiver Operating Characteristic – Area Under the Curve)}:  
The ROC-AUC represents the model’s overall ability to distinguish between pneumonia and normal cases across all classification thresholds. A higher ROC-AUC indicates stronger discriminative performance, independent of the decision threshold. 

\textbf{PR-AUC (Precision–Recall – Area Under the Curve)}:  
The PR-AUC summarizes the trade-off between precision and recall across all thresholds. It is particularly informative in imbalanced datasets because it emphasizes the model’s ability to correctly detect the minority class.

\subsection{Pneumonia Localization}
To highlight the most influential regions in predictions, we employ Gradient weighted Class Activation Mapping (Grad-CAM) \cite{Selvaraju_2017} serving as a weakly supervised localization mechanism and enhancing clinical interpretability.  
For CNN based architectures, Grad-CAM is computed using the feature maps and gradients of the last convolution layer, which provide a direct spatial correspondence with the input image. Whereas, ViT operate on patch embeddings instead of convolution features, therefore we extend Grad-CAM formulation by capturing activations and gradients from the final MLP block of the last transformer encoder layer.

\subsubsection{Grad-CAM for CNN Architectures}

In convolutional architectures such as ResNet, DenseNet, and MobileNet, Grad-CAM is applied to the \textit{last convolutional layer}, which retains the highest-level semantic and spatial information. 
Let $A \in \mathbb{R}^{C \times H \times W}$ denote the activation maps of this layer, and $\nabla Y_c \in \mathbb{R}^{C \times H \times W}$ represent the gradients of the predicted class score $Y_c$ with respect to these activations. 
The channel-wise importance weights are obtained by global average pooling of the gradients:

\begin{equation}
\alpha_k = \frac{1}{H W} \sum_{i=1}^{H}\sum_{j=1}^{W}\nabla Y_c[k,i,j],
\end{equation}

and the class-discriminative heatmap is computed as:

\begin{equation}
\text{CAM}(i,j) = \text{ReLU}\left( \sum_{k=1}^{C} \alpha_k A_k(i,j) \right).
\end{equation}

The resulting activation map is upsampled to match the original image resolution and combined with the input image to generate a heatmap overlay that highlights the regions most responsible for the model’s decision.

\subsubsection{Grad-CAM for Vision Transformers}

Vision Transformers (ViTs) replace convolutional filters with tokenized patch embeddings, requiring a modification of the Grad-CAM formulation. 
We capture activations from the \textit{final Linear layer of the last MLP block} within the last transformer encoder, which preserves spatially meaningful representations for all image patches.
Let $A \in \mathbb{R}^{N \times C}$ be the activations of $N$ patch tokens (excluding the class token) and $\nabla Y_c \in \mathbb{R}^{N \times C}$ the corresponding gradients of the predicted class. 
The importance weights are computed as:

\begin{equation}
\alpha_k = \frac{1}{N} \sum_{i=1}^{N} \nabla Y_c[i,k],
\end{equation}

and the patch-level class activation map is obtained as:

\begin{equation}
\text{CAM}(i) = \text{ReLU}\left( \sum_{k=1}^{C} \alpha_k A[i,k] \right).
\end{equation}

The one-dimensional patch map is reshaped into a 2D grid $(H_p \times W_p)$ based on the number of patches and subsequently upsampled to the input image resolution. 
The resulting heatmap is overlaid on the original image to visualize the spatial contribution of each patch to the prediction.

\noindent
This unified Grad-CAM framework provides consistent visual interpretability across both CNN and transformer-based backbones, enabling qualitative comparison of their attention on diagnostically relevant regions.

\subsection{Quantitative Localization Evaluation}
Since pixel-level annotations are not available in chest X-ray dataset, we adopt a lightweight quantitative metric to evaluate the anatomical consistency of Grad-CAM explanations. Specifically, we compute a Lung Attention Ratio (LAR), defined as the proportion of Grad-CAM activation energy that falls within a coarse lung region of interest (ROI).

The lung ROI is defined using a fixed thoracic anatomical prior that excludes image borders and sub-diaphragmatic regions. This ROI does not represent precise lung segmentation and is used solely for evaluation purposes. For each input image, Grad-CAM heatmaps are normalized and only the top 20\% of activation values are retained to suppress background noise. LAR is then computed as the ratio of activation within the lung ROI and the total activation across the image, as shown in Eq (\ref{eq:lar}).

Quantitative evaluation is performed on a fixed representative subset of test images from each class, and the same subset is used across all evaluated architectures.
\begin{equation}
\label{eq:lar}
\mathrm{LAR} =
\frac{
\sum_{(x,y)\in \Omega_{\mathrm{lung}}} A(x,y)
}{
\sum_{(x,y)} A(x,y)
}
\end{equation}

where $A(x,y)$ denotes the Grad-CAM activation at spatial location $(x,y)$,  $\Omega_{\mathrm{lung}}$ represents the coarse lung region of interest, and $\Omega$ denotes the full image domain.

\begin{figure}[htbp]
\centerline{\includegraphics[width=\linewidth]{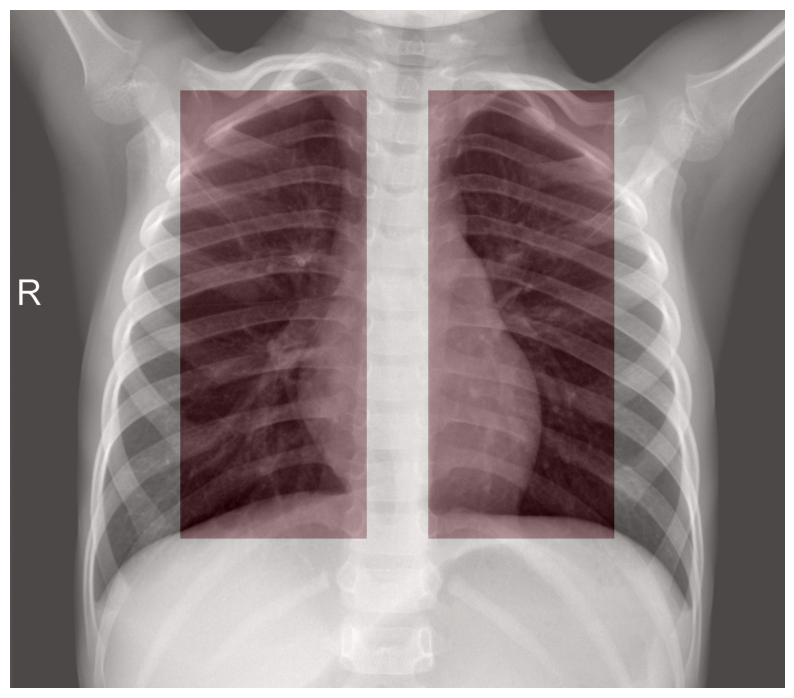}}
\caption{Illustration of the coarse lung region of interest (ROI) used for quantitative localization evaluation.}
\label{fig:lung_roi}
\end{figure}

\section{Experiments}
The details of the experiments, including the datasets, loss functions, model training, results and analysis are described as follows:

\subsection{Experimental Setup}
All models were trained under identical conditions to ensure fair comparison. Training and evaluation were performed using PyTorch 2.8.0+cu126 on an NVIDIA T4 GPU with 15 GB VRAM.  The batch size was 32, learning rate $1\times10^{-4}$, and weight decay $1\times10^{-4}$.  Each model was trained for 10 epochs with early stopping based on validation ROC-AUC. The best checkpoint per backbone was saved and later evaluated on the independent test split. Evaluation metrics include Accuracy, ROC-AUC, PR-AUC and Best F1.

\subsection{Result}

\begin{table*}[htbp]
\centering
\caption{Performance comparison among the evaluated architectures on the Chest X-Rays dataset~\cite{KERMANY2018}.}
\label{tab:performance}
\setlength{\tabcolsep}{6pt}
\renewcommand{\arraystretch}{1.1}
\footnotesize
\begin{tabular}{lcccccc}
\hline
\textbf{Model} & \textbf{Val Acc} & \textbf{Test Acc} & \textbf{ROC-AUC} & \textbf{PR-AUC} & \textbf{F1} & \textbf{Params (M)} \\
\hline
ResNet-18       & 97.5\% & 98\% & 0.9971 & 0.9990 & 0.987 & 11.5 \\
ResNet-50       & 96.8\% & 96\% & 0.9952 & 0.9983 & 0.981 & 24.6 \\
DenseNet-121    & 97.9\% & 97\% & 0.9955 & 0.9984 & 0.984 & 7.5 \\
EfficientNet-B0 & 96.9\% & 98\% & 0.9971 & 0.9989 & 0.987 & 4.7 \\
MobileNet-V2    & 95.6\% & 97\% & 0.9946 & 0.9980 & 0.982 & 2.9 \\
MobileNet-V3    & 96.2\% & 97\% & 0.9971 & 0.9990 & 0.987 & 4.9 \\
ViT             & 96.2\% & 97\% & 0.9971 & 0.9990 & 0.987 & 86.2 \\
\hline
\end{tabular}
\end{table*}

\begin{table*}[ht!]
\centering
\caption{Per class quantitative evaluation of performance (Precision, Recall, Specificity) for all evaluated architectures on the Chest X-Ray test set~\cite{KERMANY2018}.}
\label{tab:per_class_metrics}
\begin{tabular}{lcccc}
\hline
\textbf{Model} & \textbf{Class} & Precision & Recall & Specificity \\
\hline
ResNet-18 & Normal     & 0.97 & 0.96 & 0.994 \\
          & Pneumonia  & 0.99 & 0.99 & 0.958 \\
\hline
ResNet-50 & Normal     & 0.93 & 0.94 & 0.993 \\
          & Pneumonia  & 0.98 & 0.97 & 0.966 \\
\hline
DenseNet-121 & Normal    & 0.97 & 0.93 & 0.928 \\
             & Pneumonia & 0.99 & 0.96 & 0.991 \\
\hline
EfficientNet-B0 & Normal    & 0.95 & 0.97 & 0.966 \\
                & Pneumonia & 0.99 & 0.98 & 0.983 \\
\hline
MobileNet-V2 & Normal    & 0.94 & 0.95 & 0.993 \\
             & Pneumonia & 0.98 & 0.98 & 0.966 \\
\hline
MobileNet-V3 & Normal    & 0.93 & 0.97 & 0.970 \\
             & Pneumonia & 0.99 & 0.97 & 0.975 \\
\hline
ViT & Normal    & 0.95 & 0.95 & 0.953 \\
    & Pneumonia & 0.98 & 0.98 & 0.981 \\
\hline
\end{tabular}
\end{table*}

Overall, all evaluated architectures achieved strong discriminative performance on the pneumonia classification task, with test accuracies ranging between 96–98\%. Among them, ResNet-18 and EfficientNet-B0 achieved the highest test accuracy of 98\% with an F1-score of 0.987, while maintaining ROC-AUC and PR-AUC values above 0.997. Despite its smaller size, MobileNet-V3 Large delivered comparable accuracy of 97\%, demonstrating its suitability for mobile and embedded clinical applications. In contrast, deeper backbones such as ResNet-50 and DenseNet-121 exhibited marginally lower generalization performance, suggesting mild overfitting. These results indicate that compact architectures, when combined with focal loss and patient-wise splitting, can achieve high diagnostic accuracy while remaining computationally efficient.

\begin{figure*}[htbp]
\centering

\begin{subfigure}{\textwidth}
  \centering
  \includegraphics[width=\textwidth]{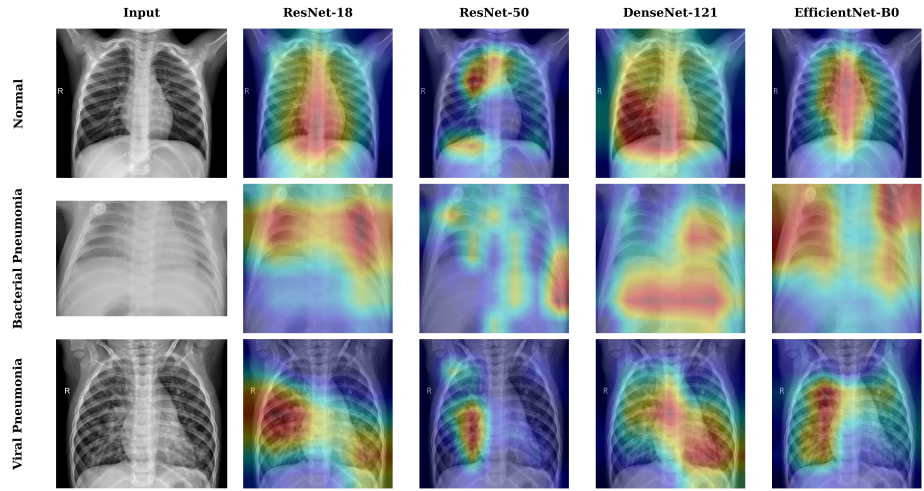}
  \caption{ResNet / DenseNet / EfficientNet.}
\end{subfigure}

\vspace{2mm}

\begin{subfigure}{\textwidth}
  \centering
  \includegraphics[width=\textwidth]{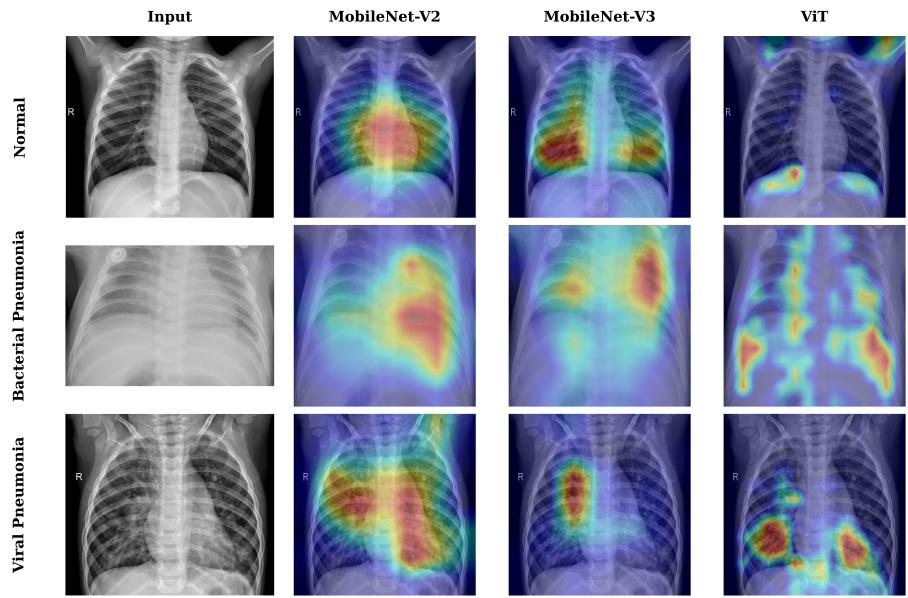}
  \caption{MobileNet and ViT.}
\end{subfigure}

\caption{Grad-CAM overlays for a normal chest X-ray and two pneumonia cases (bacterial, viral) across seven backbones. Bright regions indicate strong model attention toward pneumonia related features.}
\label{fig:gradcam}
\end{figure*}

FIGURE~\ref{fig:gradcam} illustrates Grad-CAM overlays for representative Normal, Bacterial, and Viral Pneumonia samples across the evaluated architectures. For Normal chest X-rays, activation responses are generally weak and spatially diffuse, often extending beyond lung boundaries, indicating low diagnostic confidence in the absence of pathology. An exception is DenseNet-121, which exhibits spurious activation along the left lung region, suggesting mild sensitivity to background intensity variations or residual noise.

\begin{table*}[h!]
\centering
\caption{Quantitative Grad-CAM localization using Lung Attention Ratio (LAR) on a representative subset of the test set.}
\label{tab:localization_metrics}
\begin{tabular}{lcccc}
\hline
\textbf{Model} & \textbf{Normal} & \textbf{Bacterial Pneumonia} & \textbf{Viral Pneumonia} \\
\hline
ResNet-18      & $0.547 \pm 0.157$ & $0.242 \pm 0.137$ & $0.375 \pm 0.158$ \\
ResNet-50      & $0.593 \pm 0.119$ & $0.320 \pm 0.086$ & $0.272 \pm 0.190$ \\
DenseNet-121   & $0.448 \pm 0.241$ & $0.417 \pm 0.128$ & $0.429 \pm 0.143$ \\
EfficientNet-B0& $0.589 \pm 0.163$ & $0.343 \pm 0.163$ & $0.225 \pm 0.214$ \\
MobileNet-V2   & $0.612 \pm 0.128$ & $0.410 \pm 0.201$ & $0.409 \pm 0.180$ \\
MobileNet-V3   & $0.553 \pm 0.184$ & $\mathbf{0.693 \pm 0.072}$ & $\mathbf{0.584 \pm 0.280}$ \\
ViT-B/16       & $\mathbf{0.172 \pm 0.038}$ & $0.605 \pm 0.133$ & $0.381 \pm 0.096$ \\
\hline
\end{tabular}
\end{table*}

In contrast, pneumonia cases produce focused and high-intensity activations within pulmonary regions corresponding to radiographic opacities, particularly in the middle and lower lung zones. Among the CNN backbones, MobileNet-V3 produces the most compact and noise-free localization across classes, while ResNet-18 and DenseNet-121 also demonstrate well-defined activations for pneumonia cases. Although EfficientNet-B0 achieves high classification accuracy (98\%), its Grad-CAM visualizations are comparatively diffuse and occasionally midline-biased. Similarly, ResNet-50 displays intermittent off-target hotspots.

Quantitative localization results are summarized in TABLE~\ref{tab:localization_metrics}. MobileNet-V3 achieves stable lung-focused attention for pneumonia cases, with a Lung Attention Ratio (LAR) of $0.693$ with deviation of $0.072$ for Bacterial Pneumonia, indicating low variance and consistent localization behavior. In contrast, ViT-B/16 exhibits clearer discrimination between Normal and Bacterial Pneumonia samples, with a substantially lower LAR for Normal images $0.172$ with deviation of $0.038$ and higher LAR for Bacterial Pneumonia $0.605$. However, its separation for Viral Pneumonia is less pronounced $0.381$ with deviation of $\pm 0.096$, reflecting broader and more diffuse attention patterns associated with global self-attention.

Overall, the combined qualitative and quantitative analyses demonstrate that the proposed models predominantly attend to clinically meaningful lung regions. In particular, MobileNet-V3 achieves a favorable balance between localization stability, interpretability, and computational efficiency, reinforcing its potential for trustworthy and deployable AI-assisted pneumonia screening.

\section{Conclusion and Future Works}
This study benchmarked multiple CNN backbones and Vision Transformer for weakly supervised pneumonia localization using only image level supervision. All models achieved high discriminative performance, with test accuracies ranging from 96\% to 98\%. ResNet-18 and EfficientNet-B0 consistently outperformed deeper networks, demonstrating that compact architectures can generalize well when trained with class-balanced sampling and focal loss. Grad-CAM heatmaps confirmed that attention focused on radiologically relevant opacities, validating interpretability and trustworthiness. The results further show that lightweight models, such as MobileNet-V3, can deliver near state-of-the-art (SOTA) accuracy with low computational cost, facilitating edge device or mobile health deployments.

Although the proposed framework demonstrates the effectiveness of Grad-CAM based weakly supervised localization for pneumonia detection, several opportunities for extension remain open for future investigations, as outlined below.
\begin{itemize}
    \item This research is currently limited to a single dataset. Further work should involve evaluation on larger and more diverse datasets, such as RSNA Pneumonia and NIH ChestX-ray14, to enhance robustness and generalization. 
    \item  Further extensions may explore multi-label thoracic disease localization, radiologist reader studies, and mobile deployment optimizations to strengthen 11 the framework’s clinical relevance and translational impact.
\end{itemize}

Overall, this study highlights that explainable and weakly supervised deep-learning methods can bridge the gap between black-box image classification and clinically interpretable decision support for pneumonia detection.

\vspace{5mm}
\noindent\textbf{Author Contributions}\\
K.S.: conceptualization, methodology, software; K.S. and A.B.: data curation, writing—original draft preparation; K.S. and A.B: visualization, investigation; K.S.: supervision; K.S. and A.B.: software, validation; writing—reviewing and editing. All authors have read and agreed to the published version of the manuscript.

\vspace{5mm}
\noindent\textbf{Funding}\\
This research received no external funding

\vspace{5mm}
\noindent\textbf{Institutional Review Board Statement}\\
Not applicable

\vspace{5mm}
\noindent\textbf{Informed Consent Statement}\\
Not applicable

\vspace{5mm}
\noindent\textbf{Data Availability Statement}\\
The dataset used in this study is publicly available (see Reference~\cite{KERMANY2018} for the Kermany dataset). 
The source code supporting the findings of this study is available at: 
\url{https://github.com/kiranshahi/pneumonia-analysis}.

\vspace{5mm}
\noindent\textbf{Acknowledgments}\\
The author acknowledges the use of the Chest X-ray dataset by Kermany et al. \cite{KERMANY2018}.

\vspace{5mm}
\noindent\textbf{Conflict of Interest Disclosure}\\
The author declares no conflict of interest.

\bibliographystyle{unsrt}
\bibliography{main}

\end{document}